%% file: main.tex
\documentclass[preprint,12pt]{elsarticle}

\usepackage{amsmath}
\usepackage{amssymb}
\usepackage{graphicx} % Load only once
\usepackage{lipsum}
\usepackage{algorithm}
\usepackage{algpseudocode}
\usepackage{booktabs}
\usepackage{array}
\usepackage{tabularx}
\usepackage{xurl}
\usepackage{xr-hyper}
\usepackage{custom-style}
\renewcommand{\covercorrespondence}{\vspace{1cm}\textcolor{black}{Corresponding author:}\enspace\href{mailto:eftim.zdravevski@magix.ai}{\texttt{eftim.zdravevski@magix.ai}}}
\usepackage{enumitem}
\usepackage{times}


\newpageafter{author}

\begin{document}

\begin{frontmatter}

\title{Federated versus Centralized PV Capacity-Factor Forecasting:
A Non-Inferiority Study Across Seven Stations in North Macedonia}

\author[1]{Amar Kumar Mandal}

\author[2]{Yuvraj Chaudhary}

\author[3,4]{Dimitar Kitanovski}

\author[3,4]{Eftim Zdravevski}

\affiliation[1]{
  organization={Kadir Has University},
  city={Istanbul},
  country={Turkey}
}

\affiliation[2]{
  organization={Department of Chemical and Biological Engineering,
  Friedrich Alexander University Erlangen-Nuremberg},
  city={91058 Erlangen},
  country={Germany}
}

\affiliation[3]{
  organization={Faculty of Computer Science and Engineering,
  Ss. Cyril and Methodius University},
  addressline={Rugjer Boshkovik 16},
  city={Skopje},
  country={North Macedonia}
}

\affiliation[4]{
  organization={Magix.AI},
  addressline={Vasil Gjorgov 19/1-3},
  city={Skopje},
  country={North Macedonia}
}

\begin{abstract}
\textbf{Background:}
Photovoltaic (PV) forecasts inform energy commitments by PV operators and aggregators and guide charge–discharge scheduling in solar-plus-storage systems. Federated learning offers a framework for collaborative model training through the exchange of model updates in an environment where local operators cannot pool raw operational data. Its suitability for such collaboration depends in part on whether it can retain forecasting accuracy comparable to centralized training. This study evaluated whether federated training was non-inferior to centralized training for PV capacity-factor prediction across seven stations in North Macedonia. A common multilayer perceptron architecture was used in an offline forecasting comparison with a 24-hour timestamp lag between historical inputs and target intervals. Exploratory validation compared local, centralized, Federated Averaging, and Federated Proximal (FedProx) training. FedProx ($E = 1$, $\mu = 1$) was selected for a confirmatory non-inferiority comparison under a protocol fixed before test evaluation. This comparison used past-only offline inputs, five paired training seeds, macro-averaged mean absolute error (MAE), a pre-specified non-inferiority margin of 0.005, and a paired day bootstrap with 10,000 repetitions over 49 test dates.

\par\noindent\textbf{Results:}
FedProx achieved a macro-MAE of 0.128397, compared with 0.133405 for centralized training. The paired difference (FedProx minus centralized) was $-0.005008$ (95\% confidence interval: $[-0.009926, -0.000537]$). The upper confidence bound was below the pre-specified margin, satisfying the non-inferiority criterion. All five seed-specific differences favored FedProx, and its mean MAE was lower at six of the seven stations. Post hoc calendar sensitivity analyses did not change the conclusion.

\par\noindent\textbf{Conclusions:}
These findings support further evaluation of federated forecasting for collaboration under data-pooling constraints. The study therefore demonstrates predictive non-inferiority within the evaluated setting. Benefits for energy commitments and battery scheduling remain potential applications requiring operational evaluation.
\end{abstract}


\end{frontmatter}

\clearpage
\newpage
\newpage
\input{sections/introduction}
\input{sections/methodology}
\input{sections/results}
\input{sections/conclusion}

\section*{Supplementary Material}
The accompanying supplementary material provides supporting tables, the MLP
architecture diagram, the FedAvg/FedProx algorithm, secondary
validation comparisons, margin sensitivity, and evaluation metric definitions.

\section*{Declarations}

\subsection*{Ethics approval and consent to participate}
Not applicable. This study analyzed photovoltaic production and meteorological data and did not involve human participants, human data or tissue, or animals.

\subsection*{Consent for publication}
Not applicable. This manuscript does not contain any individual person’s data, images, or videos.

\subsection*{Availability of data and materials}
The SolarMagix source data used in this study are publicly available through the
original dataset publication \cite{Daskalov2026}. The source code and accompanying
assets required to reproduce the preprocessing, model training, evaluation,
statistical analysis, and figures are archived on Zenodo as \emph{SolarFL: Federated
PV Forecasting}, version v1.1.0, with DOI
\href{https://doi.org/10.5281/zenodo.22460184}{10.5281/zenodo.22460184}
\cite{PVForecastingRepository2026}.

\subsection*{Competing interests}
The authors declare that they have no known competing interests or personal
relationships that could have appeared to influence the work reported in this paper.

\subsection*{Funding}
This work was partially funded by the Faculty of Computer Science and Engineering,
Ss. Cyril and Methodius University in Skopje. It was also partially financed by the
Ministry of Education and Science of the Republic of North Macedonia through the
project ``Utilising AI and National Large Language Models to Advance Macedonian
Language Capabilities.'' This work was co-funded by the European Union under Grant
Agreement No. 101263128 (VEZILKA). The work presented in this article was also
partially funded by Magix.AI, Skopje, North Macedonia.

\subsection*{Authors' contributions}
\textbf{Amar Mandal}: Conceptualization, Methodology, Software, Formal Analysis, Data
Curation, Writing--Original Draft, Visualization. \textbf{Yuvraj Chaudhary}:
Conceptualization, Methodology, Formal Analysis, Writing--Original Draft. \textbf{Dimitar Kitanovski}: Conceptualization, Methodology, Validation, Resources, Writing--Review \& Editing. 
\textbf{Eftim Zdravevski}: Conceptualization, Writing--Review \& Editing,
Supervision, Funding Acquisition. 

\subsection*{Acknowledgements}
Not applicable.

\newpage
\bibliographystyle{elsarticle-num}
\setlength{\emergencystretch}{2em}
\hbadness=2000
\bibliography{references}

\end{document}

\begin{center}
\textbf{Supplementary Material}\par
\medskip
Federated versus Centralized PV Capacity-Factor Forecasting:\\
A Non-Inferiority Study Across Seven Stations in North Macedonia\par
\medskip
Amar Kumar Mandal, Yuvraj Chaudhary, Dimitar Kitanovski, and Eftim Zdravevski
\end{center}

Station labels S1--S7 refer to the seven station clients throughout.
The S-prefixed numbering of supplementary sections, tables, figures, algorithms,
and equations is independent of these station labels.

\section{Experimental Evidence Scope and Split Boundaries}
\label{supp:evidence_scope}
\input{sections/experiment_scope_table}

\begin{table}[H]
\centering
\small
\setlength{\tabcolsep}{3pt}
\renewcommand{\arraystretch}{1.12}
\caption{Frozen chronological split boundaries (dates only) and raw row counts before
feature eligibility filtering. Full UTC boundaries and retained counts are supplied in
\texttt{split\_counts.csv}. S6's last raw training interval was 28 April at 18:00.
Its validation started on 29 April.}
\label{tab:split_boundaries}
\begin{tabular}{@{}l c c c c r@{}}
\toprule
\textbf{Client} & \textbf{Train from} & \textbf{Train to} & \textbf{Val. to} &
\textbf{Test to} & \textbf{Rows (train/val./test)} \\
\midrule
S1 & 2024-12-31 & 2025-04-17 & 2025-05-14 & 2025-05-31 & 1{,}729 / 370 / 371 \\
S2 & 2025-01-01 & 2025-05-01 & 2025-05-16 & 2025-05-31 & 1{,}138 / 244 / 245 \\
S3 & 2024-12-31 & 2025-04-28 & 2025-05-15 & 2025-05-31 & 1{,}252 / 268 / 269 \\
S4 & 2024-12-31 & 2025-04-28 & 2025-05-15 & 2025-05-31 & 1{,}238 / 265 / 266 \\
S5 & 2024-12-31 & 2025-04-28 & 2025-05-15 & 2025-05-31 & 1{,}239 / 265 / 266 \\
S6 & 2024-12-31 & 2025-04-28 & 2025-05-15 & 2025-05-31 & 1{,}231 / 264 / 264 \\
S7 & 2024-06-27 & 2025-02-19 & 2025-04-13 & 2025-05-31 & 4{,}759 / 1{,}020 / 1{,}020 \\
\bottomrule
\end{tabular}
\end{table}

\clearpage
\section{Feature Dimensions and Optimization Protocol}
\label{supp:model_configuration}
\input{sections/feature_dimensions_table}
\input{sections/optimization_table}

\clearpage
\section{Model Architecture and Federated Training Algorithm}
\label{supp:training_algorithm}

\begin{figure}[H]
    \centering
    \includegraphics[width=\textwidth]{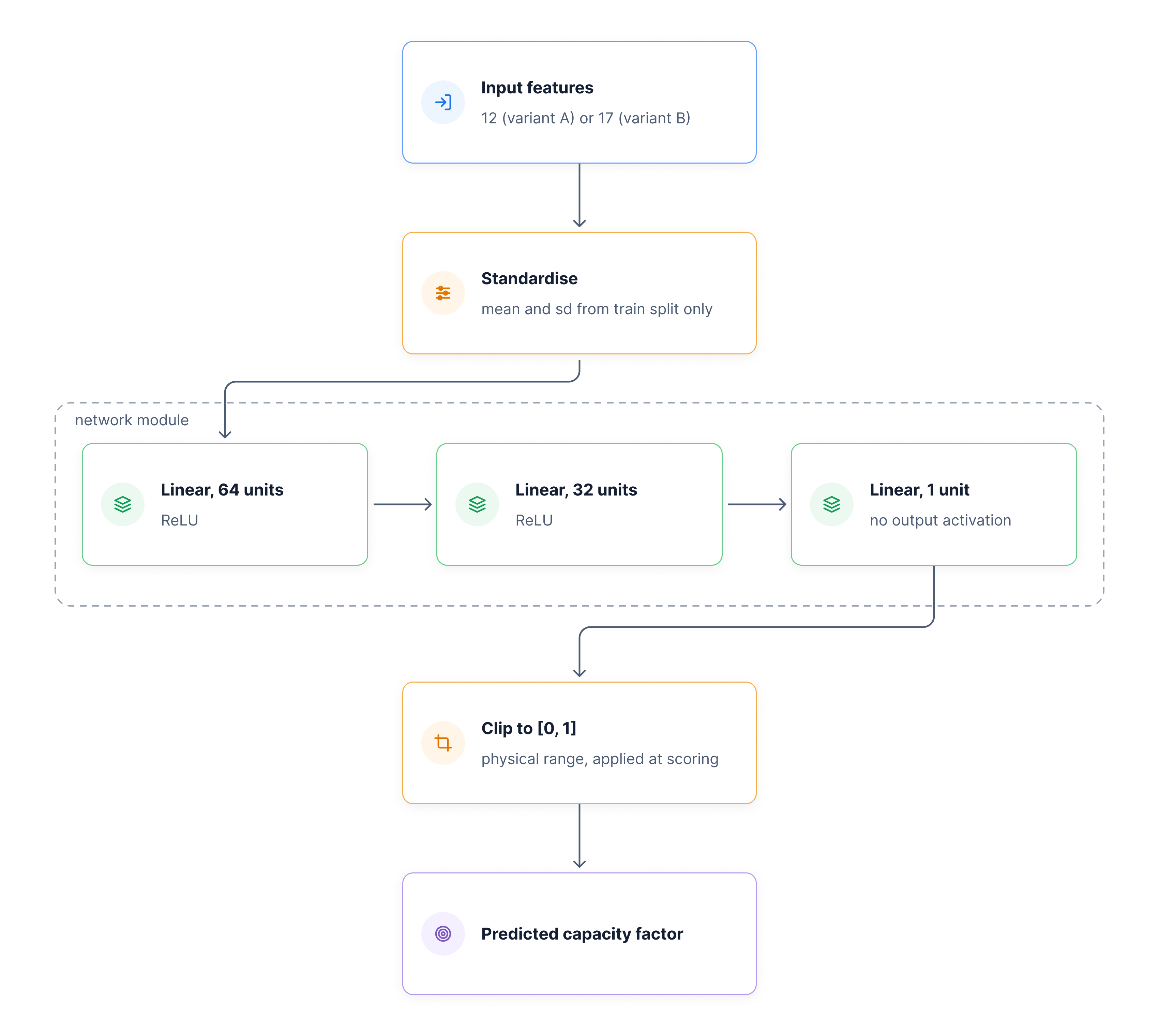}
    \caption{Shared MLP: input widths 12/17 for A/B, with 11/16 pooled nonconstant
    columns and 2,945/3,265 parameters. Hidden layers have 64 and 32 units with ReLU,
    followed by a linear output. Variant B adds non-operational target-hour reanalysis. The illustrated
    clipping applies to test scoring, not validation early stopping. Sharing an
    architecture does not eliminate optimizer-state or aggregation differences.}
    \label{fig:model_architecture}
\end{figure}

\begin{algorithm}[H]
\caption{FedAvg / FedProx (Full Participation, One Fit with Seed $s$)}
\label{alg:fedavg_fedprox}

\begin{algorithmic}
\Require Client training arrays $\{(\mathbf{Z}_k,\mathbf{y}_k)\}_{k=1}^{K}$ with retained counts $\{n_k\}_{k=1}^{K}$,
pooled validation arrays, training seed $s$,
local epochs $E$, proximal strength $\mu$ ($\mu=0$ recovers FedAvg),
maximum rounds $R=300$, patience $P=25$

\State Build the scaler from training-only $(\sum x,\sum x^2,n)$
\State Apply it to each client's training arrays and the pooled validation arrays
\State Initialize $\mathbf{w}^{0}$ with seed $s$; best MAE $\gets\infty$; stale rounds $\gets0$
\State Initialize persistent client shuffle streams with seeds $s+k-1$

\For{$t=0,\ldots,R-1$}

    \For{$k=1,\ldots,K$}

        \State Initialize a fresh Adam optimizer ($\eta=10^{-3}$; nominal batch size $256$)
        \State $\mathbf{w}_k \gets \mathbf{w}^{t}$

        \For{$e=1,\ldots,E$}
            \State Shuffle using the client's stream; partition with \texttt{array\_split}
            \State Update $\mathbf{w}_k$ by minimizing the mini-batch objective
            \[
            \mathcal{L}_k(\mathbf{w}_k)
            =
            \mathrm{MSE}_k(\mathbf{w}_k)
            +
            \frac{\mu}{2}
            \left\|
            \mathbf{w}_k-\mathbf{w}^{t}
            \right\|_2^2
            \]
        \EndFor

    \EndFor

    \State Aggregate
    \[
    \mathbf{w}^{t+1}
    =
    \sum_{k=1}^{K}
    \frac{n_k}{\sum_{j=1}^{K}n_j}
    \mathbf{w}_k
    \]

    \State Evaluate pooled validation MAE on unclipped predictions
    \If{MAE is strictly below the best MAE}
        \State Save $\mathbf{w}^{t+1}$ and its MAE; stale rounds $\gets0$
    \Else
        \State stale rounds $\gets$ stale rounds $+1$
    \EndIf

    \If{stale rounds $\geq P$}
        \State \textbf{break}
    \EndIf

\EndFor

\State Restore and return the best global weights, including at the round limit

\end{algorithmic}
\end{algorithm}

\section{Secondary Federated-Learning Findings}
\label{supp:secondary_fl}

The comparisons in this section are exploratory seed-0 validation results, not frozen test endpoints. For the past-only
feature variant (Variant A), FedProx $E = 1$ had lower macro-MAE than FedAvg $E = 1$
(0.116 versus 0.119). When the local training epoch ($E$) increases from $E = 1$ to
$E = 5$, the performance of both models improved, and the macro-MAE reduced from 0.119
to 0.116 for FedAvg and from 0.116 to 0.114 for FedProx. Additionally, under the same
local-epoch setting $E = 5$, FedProx had a lower MAE than FedAvg (0.114 versus 0.116).
These results are consistent with additional local training and proximal regularization
affecting optimization, but validation results alone are insufficient to prove which
mechanism caused the changes.

Increasing the number of local epochs ($E$) allows each client to train longer between server aggregations. However, this alone does not establish communication efficiency. The $E = 5$ configurations used approximately five times as much local training per communication round as $E = 1$ and were not compute-matched to centralized training. These exploratory comparisons therefore do not establish comparative computational or communication efficiency. The $E = 5$ and FedAvg results remain outside the pre-specified confirmatory comparison and do not support the main non-inferiority conclusion under its margin of 0.005.

\input{sections/appendix}

%% file: sections/introduction.tex
\section{Introduction}

PV has emerged as one of the most promising renewable energy technologies worldwide. Annual PV installations are growing exponentially, reaching nearly 700 GW in 2025 and nearing 1 TW in annual capacity \cite{IEA2026}. Solar intermittency still challenges technology penetration and reliability by inducing weather-driven fluctuations, grid instability, reserve requirements, etc. Additionally, incoming solar radiation is expected to be altered in future climates due to greenhouse gases, aerosols, and cloud cover, which complicates long-term planning \cite{Yin2020,Boucher2013Clouds}. This underscores the need for PV power forecasts in complex meteorological conditions. Forecasting relies on either direct power predictions or indirect methods that first model solar irradiance before converting it to power \cite{ANTONANZAS201678}. In radiometric terms, solar irradiance is the instantaneous radiant power incident per unit area (W/m$^2$), whereas solar irradiation is the radiant energy per unit area integrated over a specified time interval (J/m$^2$ or Wh/m$^2$). Accurate prediction of this parameter is important for grid integration and energy markets, but it remained highly immature in the last decade compared to other energy forecasting \cite{HONG2016896}. PV operators and aggregators in day-ahead electricity markets use generation forecasts to decide how much energy to commit for each trading interval. Forecast errors expose them to imbalance-settlement costs \cite{BASHIR2021100477}. Operators of PV plants coupled with battery energy storage systems (BESS) use forecasts to plan when and how much to charge or discharge, subject to battery energy and power limits. Forecast-informed scheduling has been investigated in hybrid PV--wind plants with storage, linking generation predictions to operating decisions \cite{YANG20192326}. These applications motivate reliable forecasting, although operational value depends on the decision strategy and operating conditions as well as prediction accuracy.

Solar-forecasting methods relied on physical and statistical approaches earlier. Physical methods used solar geometry, clear-sky radiation models, atmospheric radiative-transfer models, and numerical weather prediction (NWP) to describe the influence of meteorological conditions on incoming radiation. Statistical approaches extracted temporal relationships from historical irradiance or PV-power measurements using persistence models, autoregressive models, regression methods, and exponential smoothing techniques \cite{INMAN2013535,YANG201860}. Although these methods provided considerable accuracy, their reliability is often limited under rapidly changing cloud conditions and nonlinear relationships between climate parameters and PV output. Further, machine-learning-based models were subsequently adopted to model the nonlinear relationship between solar radiation, meteorological variables, and PV output. Approaches based on artificial neural networks, support-vector regression, fuzzy systems, decision trees, and ensemble learning demonstrated improved reliability compared with conventional statistical models, particularly for short-term forecasting \cite{ANTONANZAS201678,VOYANT2017569,AHMED2020109792}. The development of deep learning further expanded this research direction by enabling the automatic extraction of temporal and spatial features from large and heterogeneous datasets. Recurrent neural networks (RNN), long short-term memory networks (LSTM), convolutional neural networks (CNN), and hybrid CNN--LSTM architectures have been increasingly reported for solar forecasting, using historical time series, NWP data, satellite observations, and total-sky images as inputs \cite{WANG2019111799,Rajagukguk13246623}. More recent studies have focused on hybrid and ensemble frameworks that combine physical knowledge with data-driven models, as well as on probabilistic forecasting approaches that quantify the uncertainty associated with future PV generation \cite{VANDERMEER20181484,AHMED2020109792}. This evolution reflects a gradual transition from single-site deterministic prediction toward spatially distributed, uncertainty-aware, and operationally oriented forecasting systems \cite{YANG201860}.

Existing studies have focused on improving model architecture and input features, but less attention has been paid to integrating data configuration for solar forecasting. Recent studies have addressed signal non-stationarity through decomposition, pairing Complete Ensemble Empirical Mode Decomposition with Adaptive Noise (CEEMDAN) or variational mode decomposition with LSTM ensembles \cite{PATTNAIK2025110174,HOU20245125}, while others incorporated sky-camera or webcam imagery alongside meteorological covariates for short-horizon irradiance forecasting \cite{ANSONG20254111,SARKIS2024112600}. Some studies explored LSTM and Gated Recurrent Units (GRU) variants tailored to specific settings, including dynamic sunrise-sunset windowing with online learning \cite{11153385}, net-load forecasting in mixed solar-wind microgrids \cite{10741801}, and hybrids combining GRU with gradient boosting, swarm optimization, or model-predictive control for real-time dispatch \cite{KHAYAT2025e02884,kakeu,Souhe2024}. CNN-LSTM architectures were further extended with dimensionality reduction, physics-informed constraints, and attention-based encoder-decoder designs to better capture spatial-temporal dependencies and incorporate domain knowledge such as clearness index and shortwave radiation \cite{onal,yao,gou}. More recently, Transformer-based architectures and Reformer variants have reported better stability and applicability over recurrent baselines for both short-term and ultra-short-term forecasting \cite{9941461,https://doi.org/10.1155/2022/9421400,LIU2026109228}, and generative approaches based on Generative Adversarial Networks (GANs) and diffusion models have been proposed to produce probabilistic forecasts under complex weather conditions \cite{he,lopez}. A few studies have reported PV prediction as a reinforcement learning problem, using actor-critic or policy-gradient formulations with auxiliary outlier detection and interpretability analysis \cite{xu,ZHANG2024103830}. Luo et al. proposed a physics-constrained LSTM for improved robustness under sparse data \cite{LUO2021120240}, while Akhter et al. proposed a Salp Swarm Algorithm (SSA)-optimized RNN--LSTM and compared it with genetic algorithm (GA)- and particle swarm optimization (PSO)-based variants across three PV technologies \cite{AKHTER2022118185}. Huang et al. combined CEEMDAN decomposition with an ensemble of deep temporal convolutional networks \cite{10234639}, Abdel et al. introduced a convolutional-gated recurrent architecture with bidirectional residual blocks \cite{ABDELBASSET2021127037}, and Cortez et al. developed a correction-based pipeline refining LSTM/GRU/MLP forecasts with an LSTM autoencoder \cite{Cortez2024}. Hu et al. addressed spatial heterogeneity using a graph attention network (GAT) to select representative stations before ensemble forecasting at the regional level \cite{math12162561}. Finally, Kumar et al. proposed an asymmetric actor-critic reinforcement learning framework for irradiance forecasting across climatically diverse sites \cite{11411944}, and Wang et al. used an LSTM-based TimeGAN to generate weather-correlated PV, load, and scenario data for system planning \cite{su152015007}. While these studies collectively demonstrate progress in architecture design and cross-site model selection, none address data locality or privacy-preserving training constraints.

Applying forecasting models across multiple operators also raises questions about access to training data. Where commercial confidentiality or data-governance requirements prevent raw operational records from being pooled, federated learning offers a framework for collaboration through local model training and the exchange of model updates \cite{HOSSEINI2023116900,MICHALAKOPOULOS20254223}. This motivates evaluating whether federated training can retain predictive performance comparable to centralized learning. Studies have also applied federated learning specifically to PV and solar forecasting in response to these data-locality concerns. Zhang et al. proposed a semi-asynchronous federated CNN-LSTM with personalization for short-term solar forecasting, reporting substantial gains over fully local models \cite{ZHANG20231221}; Deng et al. extended this direction with a personalized federated GAN combining TimesNet and DeepAR to address sample scarcity and heterogeneity across distributed PV stations \cite{DENG2024101537} and Bukhari et al. combined federated learning with transfer learning in a hybrid Conv-SGRU architecture \cite{MUHAMMADSALMANBUKHARI2024100520}. These studies showed that federated learning improves local training, but also used complex, custom-built architectures, such as generative models, attention-based hybrids, and transfer-learning stages. This makes it difficult to discuss whether the improvements come from federated training itself or simply from using a more powerful model. In this regard, this study uses the simple MLP architecture across all training methods, including local, centralized, FedAvg, and FedProx, to hold architecture and primary hyperparameters fixed while comparing training regimes. Optimizer-state, aggregation, and stopping dynamics still differ; the comparison does not isolate data configuration as a sole cause.

Two studies evaluate federated PV forecasting at a comparable scale to this one, using seven station-level clients. Rayappa et al. proposed SolarTransformer, a FedProx-trained two-layer Transformer evaluated across seven heterogeneous clients with real-time ECMWF weather integration and post-training compression for edge deployment \cite{R2026101446}, and Michalakopoulos et al. compared six aggregation algorithms using a stacked BiLSTM across seven Portuguese PV plants, reporting that FedProx-style proximal aggregation outperforms both centralized and localized learning in short-term (1-day) scenarios \cite{MICHALAKOPOULOS20254223}. Both studies are close in scale and federated design to the present work, yet neither study set a statistical threshold for comparability before running the experiments. Both report performance gaps as simple numbers, such as a percentage drop in RMSE or $R^2$. Neither study tested whether that gap was small enough to count as equivalent performance, and neither compared a purely operational feature set, using only forecast-origin weather data, against one using perfect-forecast weather data from reanalysis. This comparison is central to the present work. Hosseini et al. offer the closest architectural parallel, using a multi-layer perceptron to compare federated, centralized, and localized behind-the-meter PV forecasting and explicitly framing the trade-off between accuracy and privacy \cite{HOSSEINI2023116900}. However, their comparison is again outperformance-oriented rather than equivalence-tested, and relies on a single feature configuration without examining how forecast-input quality interacts with data locality. Riedel et al. similarly compare federated (with differential privacy), centralized, and local LSTM/GRU baselines, but at a much smaller scale (three residential systems) and for grid feed-in rather than day-ahead capacity-factor prediction \cite{RIEDEL2024100452}.

Lastly, studies have shown that the benefits of federated learning in PV-related forecasting extend beyond privacy-preserving comparisons. Moradzadeh et al. use federation primarily to build a generalizable model for regions lacking historical data, emphasizing spatial transferability and cyberattack robustness rather than local-versus-pooled equivalence \cite{MORADZADEH2023697}. Zhang et al. apply federated variational Bayesian inference to probabilistic irradiance forecasting, focusing on uncertainty quantification rather than deterministic day-ahead power prediction \cite{9248598}. In the load-forecasting domain, Briggs et al. compare federated, clustered-federated, centralized, and local models, followed by a personalization step. Their results show that federated learning can narrow the performance gap with centralized training, although they do not formally test for equivalence \cite{9888131}. To fill this gap, this literature establishes that federated learning is technically viable for distributed PV and energy forecasting, and that FedAvg and FedProx based aggregation can approach centralized accuracy. The present study combines three design choices: a shared compact MLP architecture and primary hyperparameters across training regimes; a pre-specified non-inferiority test restricted to centralized MLP versus selected FedProx; and exploratory seed-0 validation comparisons of past-only and perfect-weather inputs. These choices assess predictive utility on the public seven-station SolarMagix dataset \cite{Daskalov2026} without establishing privacy protection. Energy commitment and BESS scheduling therefore provide potential application contexts for this predictive comparison. However, the present study doe not evaluate dispatch decisions or quantify operational or economic benefits.

%% file: sections/methodology.tex
\section{Materials and Methods}

\subsection{Experimental Framework}
The analysis evaluated photovoltaic (PV) capacity-factor forecasting across seven
stations in North Macedonia using a nominal 24-hour interval-label horizon. Each station
was treated as a separate client, and the dataset was partitioned by station.
Two sets of model inputs, Variant A and Variant B, were prepared from the historical dataset.
The two variants differed in the weather information
provided to the model. In Variant A, weather and PV production data from the same hour
on the previous day, labeled $T-24\,\mathrm{h}$, were combined with solar-position and
calendar features calculated for the target hour $T$.
The earlier hourly data were complete only after that hour had ended, so they were not
available exactly 24 hours before $T$. Variant A was therefore evaluated as an offline
experiment rather than as a real-time forecasting system.
In Variant B, ERA5 weather reanalysis data for the target hour were also
included. These data provided retrospective estimates of the target-hour weather and
were not available when a forecast would have been issued. Variant B was used to evaluate model performance when target-hour weather estimates were supplied as inputs. Its results therefore do not measure performance under real forecasting conditions.

The objective is to determine whether federated training across the seven station clients
achieved forecasting accuracy within a predefined tolerance of centralized training on
their combined data. Both approaches were run in a single-process simulation with access
to all station datasets. Validation inputs and targets from all stations were combined
to calculate the error used for early stopping. This setup compared forecasting accuracy rather than data isolation or privacy. Non-inferiority was tested using mean absolute error (MAE) averaged equally across stations (Section~\ref{sec:noninferiority}). The upper limit of the 95\% confidence interval for the MAE difference (federated minus centralized)
had to be below 0.005.

Exploratory seed-0 validation experiments compared four training regimes:

\begin{enumerate}
    \item \textbf{Local Training}: Each station client independently trains a dedicated model exclusively on its own local training partition without cross-site parameter sharing.
    \item \textbf{Centralized Training}: A single global model is trained directly on the pooled training partitions concatenated across all seven stations.
    \item \textbf{Federated Averaging (FedAvg)}: Each client performs local gradient updates on its private partition and transmits updated model weights to a central coordinator, which aggregates client parameters via sample-count-weighted averaging.
    \item \textbf{Federated Proximal (FedProx)}: Extends FedAvg by incorporating an $L_2$ proximal regularization penalty into each client's local objective function to constrain client drift away from the broadcast global model parameters.
\end{enumerate}

The persistence baseline predicted the capacity factor observed at the same hour on
the previous day. It was used to calculate skill scores, without a separate hypothesis
test against persistence.
Validation experiments with seed 0 compared locally and centrally trained Ridge regression
and multilayer perceptron (MLP) models, along with FedAvg and FedProx using one or five
local epochs per round ($E=1$ or $5$). These experiments used both input variants.
Variant A included 12 input features, and Variant B included 17. Each variant included
one feature with a constant value across the pooled data, leaving 11 nonconstant features
in Variant A and 16 in Variant B. Variant A experiments were repeated with seeds 0--4
for centralized MLP, FedAvg ($E=1$), and FedProx ($E=1$) to assess variation in model
performance across training seeds.
The final non-inferiority test compared only centralized MLP and the selected FedProx
model ($E=1$, $\mu=1$), using Variant A and the same five training seeds for both methods.
The test protocol was fixed before test evaluation.
Local models, Ridge regression, FedAvg, $E=5$, and Variant B were evaluated only on the
validation split, not in the final non-inferiority test.
Supplementary Table~\ref{tab:experiment_scope} summarizes the models, input variants, and training seeds
used at each evaluation stage.

The source code, configuration files, and accompanying assets used to execute this
protocol and regenerate the reported results and figures are archived in the Zenodo
reproducibility repository for this study \cite{PVForecastingRepository2026}.

\begin{figure}
    \centering
    \includegraphics[width=\textwidth,height=0.82\textheight,keepaspectratio]{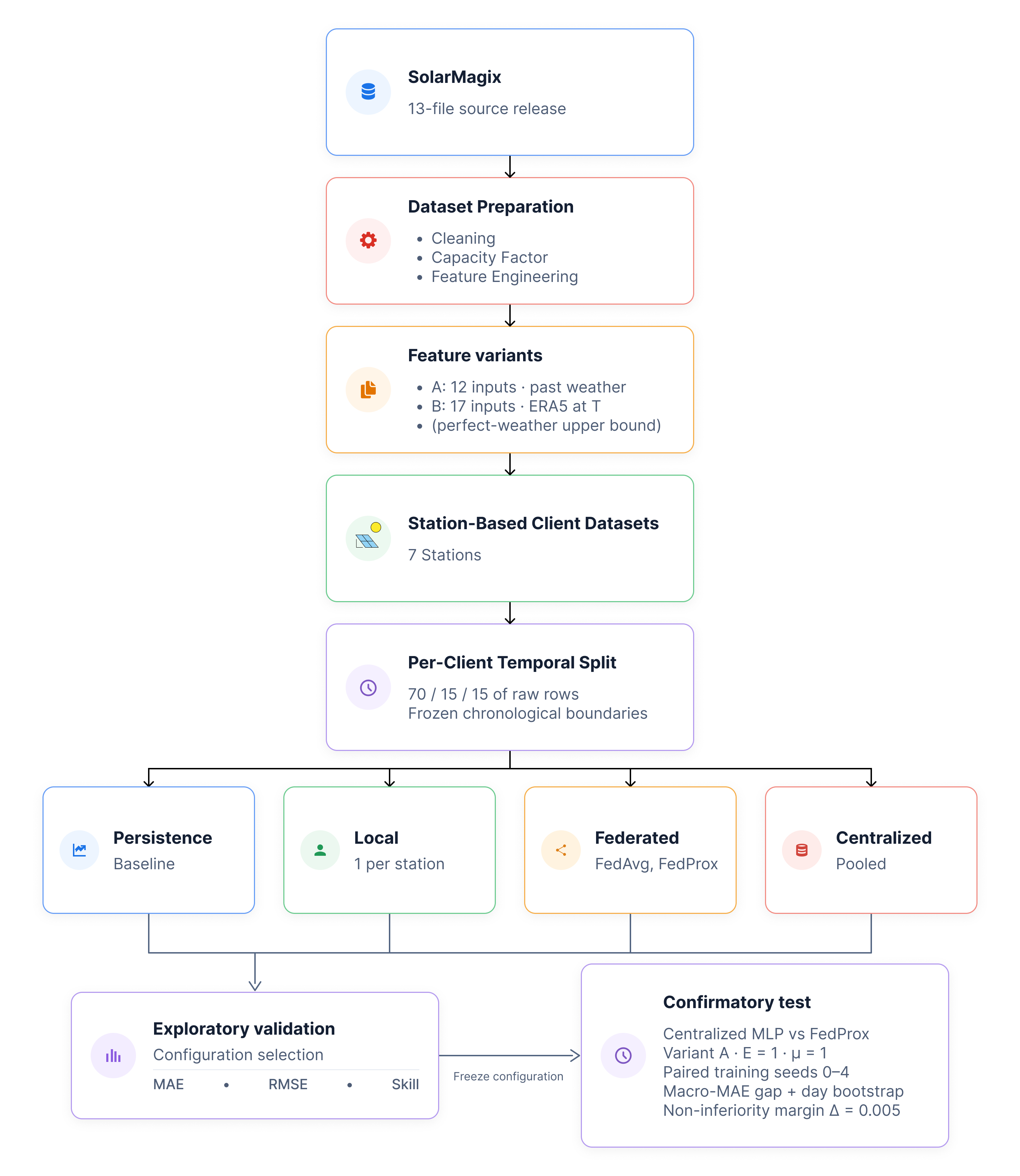}
    \caption{Evaluation flow. Broad seed-0 validation sweeps preceded the frozen
    centralized-MLP versus FedProx ($E=1$, $\mu=1$) test with paired seeds 0--4.
    Variant A used past-only inputs and a nominal 24-hour label lag, not verified
    dispatch-time availability. A/B input widths are 12/17 (11/16 pooled nonconstant
    columns). The common mask was used only for paired A/B validation. The confirmatory
    test used A and its future-independent past mask. Persistence served as the skill reference.}
    \label{fig:pipeline}
\end{figure}

\subsection{Dataset and Grain}
This study used the public SolarMagix dataset, which combines inverter telemetry from
seven stations in North Macedonia with ERA5/ERA5-Land weather reanalysis obtained through
Open-Meteo \cite{Daskalov2026}. 
The analysis used the station-level hourly records in
\path{hourly_pv_weather_station.csv}, one of the dataset's 13 source CSV files.
During data preprocessing, fixed station capacities were loaded from
\path{configs/station_labels.json} and used to calculate capacity factors.
The station capacities were recalculated from the source files to verify the stored values.
This verification used \path{devices.csv} for inverter power ratings,
\path{stations.csv} to verify each inverter's station, and
\path{hourly_pv_weather_inverter.csv} to identify inverters present in the hourly records.

The dataset contains 17{,}983 hourly observations across seven stations, spanning
27 June 2024 12:00 to 31 May 2025 23:00 UTC. Each observation represents one hour at
one station. Records for S1--S6 began on 31 December 2024, except for S2, whose records
began on 1 January 2025. Records for S7 began on 27 June 2024.
All seven stations had records through 31 May 2025. S7 accounted for about 38\% of
the observations. The analysis used separate data splits and reported model performance
for each station because the recording periods differed.

\subsubsection{Timestamp and Availability Convention}
\label{sec:timestamps}

\paragraph{Data Synchronization}
Aligning the datasets required resolving a discrepancy in how intervals were originally labeled:
\begin{itemize}
    \item \textbf{Open-Meteo Radiation Aggregates:} Hourly solar radiation averages are indexed by the \emph{end} of the averaging interval \cite{OpenMeteoHistorical2026}. For example, a 14:00 timestamp represents the data from 13:00 to 14:00.
    \item \textbf{SolarMagix PV Telemetry:} Reports hourly observations at the \emph{start} of the interval. For example, a 13:00 timestamp represents the data from 13:00 to 14:00.
\end{itemize}

Before merging, SolarMagix reassigned the preceding-hour weather aggregates to timestamps one hour earlier, separately for each station \cite{Daskalov2026}. The current study utilizes this pre-aligned dataset. Consequently, every timestamp $T$ in the finalized dataset identifies the one-hour interval starting at $T$ for both PV energy and mean irradiance. All timestamps remain in UTC without further local-time conversion.

\paragraph{Offline Evaluation vs. Real-Time Forecasting}
Variant A used PV and weather inputs from the same hour on the previous day.
Because SolarMagix labels each interval by its start time, Monday's 14:00 record
covers 14:00--15:00 and cannot be complete before 15:00. Using this record to predict
Tuesday's 14:00--15:00 production therefore leaves at most 23 hours before the target
interval starts. The study evaluated prediction accuracy offline without simulating
data-processing or ERA5 access delays. Its 24-hour lag thus describes the gap between
record timestamps, not a demonstrated ability to issue forecasts 24 hours in advance.

\subsection{Target Definition}
The regression target is the hourly capacity factor, defined in
Equation~\ref{eq:capacity_factor}, where $E_{[T,T+1\,\mathrm{h})}$ denotes the energy generated during the target hour
in kWh and $P_{\mathrm{rated}}$ denotes the station capacity in kW. The analysis
calculated each station's capacity by adding the power ratings of its included
inverters, using the values in \path{max_power}. It included only inverters with
a recorded power rating greater than zero and at least one hourly record during
the full dataset period.

\begin{equation}
\mathrm{CF}_T = \frac{E_{[T,T+1\,\mathrm{h})}}{P_{\mathrm{rated}} \cdot 1\,\mathrm{h}}
\label{eq:capacity_factor}
\end{equation}

53 out of total 66 devices had the power ratings and 50 of them met this inclusion rule. The 13 devices without ratings were excluded, where three 55~kW inverters at
S5 had no hourly records. Thus, S5's capacity factor was calculated considering the rating of  605~kW rather than 770~kW.
Table~\ref{tab:station_capacity} reports the capacities used for all seven stations.
Stations were labeled S1--S7 in ascending order of capacity, with ties resolved by
station hash.

Station capacities were calculated from inverter records across entire dataset.
The calculation did not account for when each inverter started functioning because
commissioning dates were unavailable. The absence of an inverter from the
hourly records could not verify if it was opreational or not. Therefore,
the reported capacities were derived from the dataset rather than independently
verified plant ratings. They provide a consistent basis for comparing forecasting errors
across stations, but they may not represent the capacity information that would have been
available at the time of a real-time forecast.

\begin{table}[t]
\centering
\small
\setlength{\tabcolsep}{4pt}
\caption{Capacity-denominator reconstruction from the supplied station and device audits.
Rated devices have positive, nonmissing \texttt{max\_power}; retained devices also occur
at least once in the full hourly inverter fact table. Excluded rated devices are absent
from that table, not verified uncommissioned. Unrated metadata devices are omitted.
Power sums and the final denominator are in kW.}
\label{tab:station_capacity}
\begin{tabular*}{\textwidth}{@{\extracolsep{\fill}}lrrrrrrr@{}}
\toprule
Client & Rated & Retained & Excluded & Unrated & \shortstack{All rated\\kW} & \shortstack{Excluded\\kW} & $P_{\mathrm{rated}}$ \\
\midrule
S1 & 1 & 1 & 0 & 2 & 44 & 0 & 44 \\
S2 & 1 & 1 & 0 & 1 & 44 & 0 & 44 \\
S3 & 5 & 5 & 0 & 2 & 220 & 0 & 220 \\
S4 & 8 & 8 & 0 & 2 & 440 & 0 & 440 \\
S5 & 14 & 11 & 3 & 4 & 770 & 165 & 605 \\
S6 & 11 & 11 & 0 & 2 & 605 & 0 & 605 \\
S7 & 13 & 13 & 0 & 0 & 1{,}300 & 0 & 1{,}300 \\
\midrule
Total & 53 & 50 & 3 & 13 & 3{,}423 & 165 & 3{,}258 \\
\bottomrule
\end{tabular*}
\end{table}

Capacity factors greater than one were found in 87 of the 17,983
observations, with 86 from Station S7 and 1 from Station S3. These
capacity-factor values were replaced with \texttt{NaN} instead of deleting the rows during processing. It was done in order to preserve the split manifest, which references observations by count and position.

\subsection{Feature Specification}

Input features were assembled into the model input matrix by
\texttt{build\_features()}.

\subsubsection{Feature Groups}

The input features were organized into four groups, summarized in
Table~\ref{tab:feature_groups}.

\begin{table}[t]
\centering
\small
\setlength{\tabcolsep}{6pt}
\renewcommand{\arraystretch}{1.15}
\caption{Feature groups used in the study.}
\label{tab:feature_groups}
\begin{tabular}{@{}l c c >{\raggedright\arraybackslash}p{6.2cm}@{}}
\toprule
\textbf{Group} & \textbf{Count} & \textbf{Timestamp} & \textbf{Contents} \\
\midrule
\texttt{history} & 1 & $T-24$ & Capacity factor at $T-24$ \\
\addlinespace
\texttt{geometry} & 6 & $T$ & \texttt{cos\_zenith}, \texttt{is\_daylight},
  \texttt{hour\_sin/cos}, \texttt{doy\_sin/cos} \\
\addlinespace
\texttt{weather\_past} & 5 & $T-24$ & \texttt{temperature\_2m},
  \texttt{direct\_radiation}, \texttt{diffuse\_radiation},
  \texttt{global\_tilted\_\allowbreak irradiance}, $k_t$ \\
\addlinespace
\texttt{weather\_future} & 5 & $T$ & Same five variables from target-hour ERA5 reanalysis \\
\bottomrule
\end{tabular}
\end{table}

\begin{description}[leftmargin=1.5em, style=unboxed, font=\bfseries, itemsep=0.4em]
  \item[Variant A (12 features):] Combines the \texttt{history} group (1 feature at
    $T-24\,\mathrm{h}$), the \texttt{geometry} group (6 target-hour solar-geometry and
    cyclic calendar features at $T$), and the \texttt{weather\_past} group
    (5 historical meteorological variables at $T-24\,\mathrm{h}$).

  \item[Variant B (17 features):] Appends the \texttt{weather\_future} group (the same
    5 meteorological variables extracted from retrospective ERA5 reanalysis at target
    hour $T$) to Variant A to establish a non-operational, perfect-weather performance
    ceiling.
\end{description}

\subsubsection{Derived Features}

Solar-geometry, clearness-index, and cyclic calendar features were derived from
irradiance variables and UTC timestamps.
Hourly-mean global horizontal irradiance (GHI), global tilted irradiance (GTI),
and horizontal extraterrestrial irradiance (TOA) were obtained from the dataset fields
\path{shortwave_radiation}, \path{global_tilted_irradiance}, and
\path{terrestrial_radiation}, respectively
\cite{OpenMeteoHistorical2026,OpenMeteoSolar2026}. The provider's interval-based
TOA values were used without recomputation \cite{OpenMeteoSource2026} and assigned
to $[T,T+1\,\mathrm{h})$ following the hour shift described in Section~\ref{sec:timestamps}.

The feature \texttt{cos\_zenith} was calculated from these TOA values using
Equation~\ref{eq:cos_zenith}. This feature approximates the hourly-mean cosine of
the solar zenith angle rather than an instantaneous solar-position value.
\begin{equation}
\texttt{cos\_zenith}_T =
\operatorname{clip}_{[0,1]}\!\left(
\frac{\mathrm{TOA}_T}{(1361~\mathrm{W/m^2})\,E_0(\mathrm{doy}_T)}\right).
\label{eq:cos_zenith}
\end{equation}
The calculation used a solar constant of $1361~\mathrm{W/m^2}$ and the
dimensionless Earth--Sun distance correction $E_0$, defined in
Equation~\ref{eq:earth_sun_correction}.
\begin{equation}
E_0(\mathrm{doy}) = 1 + 0.033 \cos\!\left(\frac{2\pi \cdot \mathrm{doy}}{365}\right).
\label{eq:earth_sun_correction}
\end{equation}
The day of year ($\mathrm{doy}$) was indexed from 1 using UTC timestamps, and
the calculation retained the 365-day approximation in leap years.

The dimensionless clearness index $k_t$ was calculated as the ratio of
GHI to TOA when TOA exceeded $10~\mathrm{W/m^2}$ and was assigned \texttt{NaN}
otherwise, as specified in Equation~\ref{eq:clearness_index}.
GTI was retained as a separate weather predictor.
\begin{equation}
k_t = \begin{cases}
\mathrm{GHI}/\mathrm{TOA}, & \mathrm{TOA}>10~\mathrm{W/m^2},\\
\texttt{NaN}, & \text{otherwise}.
\end{cases}
\label{eq:clearness_index}
\end{equation}
The same TOA threshold determined \texttt{is\_daylight}, which was set to 1 when
TOA exceeded $10~\mathrm{W/m^2}$ and to 0 otherwise.

Finally, the hour and day of year from UTC timestamps were encoded as the cyclic
calendar features \texttt{hour\_sin}, \texttt{hour\_cos}, \texttt{doy\_sin}, and
\texttt{doy\_cos}.

\subsubsection{Disclosed Degeneracy}

The observation sets selected by the \texttt{past} mask for Variant A and the
\texttt{common} mask for paired A/B validation both retained only daytime
observations. Daylight at the target hour was determined using deterministic
solar geometry. Consequently, the \texttt{is\_daylight} feature was constant in the
final model input matrices. Variants A and B constituted 12 and 17 input features,
respectively, of which 11 and 16 were nonconstant across the pooled observations
in each split. Supplementary Table~\ref{tab:feature_dimensions} summarizes these counts, which
do not establish the number of independent information dimensions or the matrix rank.

During standardization, the constant \texttt{is\_daylight} column was centered to
zero, and its zero standard deviation was replaced with 1. Retaining this column
preserved the frozen feature specification and model architecture, including the
parameter count. No feature-removal experiment was performed.

\subsubsection{Exclusion}

\sloppy
System-state variables, including \path{ac_power}, \path{dc_power},
\path{efficiency}, \path{device_temperature}, and
\path{perc_state_on/off/error}, were excluded to prevent direct or indirect
leakage of target capacity-factor information.
Identifiers such as \texttt{station\_hash\_id} and \texttt{source} were excluded
to avoid memorization of station identities. Globally constant features such as
\texttt{tilt} and \texttt{azimuth} were removed because they provided no variation
across observations and thus no useful information for learning.

\texttt{shortwave\_radiation} and \texttt{terrestrial\_radiation} were used to derive
features but excluded as direct model inputs. Shortwave radiation equals direct
$+$ diffuse radiation, so including it alongside these components
would introduce perfect multicollinearity without adding information.
All \texttt{*\_instant} columns were excluded because their stated interpretation
as instantaneous solar irradiance could not be validated. 
\fussy

\subsubsection{Row Set (Daylight-Only)}

Among daytime observations, the \texttt{past} filter required available Variant-A
predictors and target values, whereas \texttt{common} additionally required
the Variant-B target-hour reanalysis features. A sensitivity check confirmed that
these additional requirements excluded no otherwise eligible Variant-A observations
in this dataset. Both filters generated identical Variant-A timestamps, inputs, and
targets across the training, validation, and test splits, leaving the reported
validation and test results unchanged without retraining or numerical revision.

Table~\ref{tab:sample_counts} summarizes preprocessing across the training,
validation, and test splits. Of 17,983 raw observations, 13,398 (74.5\%) were
daytime observations. After excluding unusable daytime observations, 12,872
(71.6\% of the raw observations) remained eligible for modeling: 8,719 training,
2,059 validation, and 2,094 test observations. Table~\ref{tab:retained_split_counts}
reports raw and retained counts separately. Across clients, 95.3\% to 96.7\% of
daytime observations were retained; exclusions were primarily due to unavailable exact
24-hour lag values and invalid capacity factors. Station S7 contributed the
largest number of raw observations (6799) and had the lowest daytime retention rate
(95.3\%), consistent with its highest number of invalid capacity-factor values.

\begin{table}[t]
\centering
\small
\setlength{\tabcolsep}{8pt}
\renewcommand{\arraystretch}{1.15}
\caption{Sample counts per client across processing stages, summed over training,
validation, and test partitions. ``Model matrix'' denotes all eligible rows, not training alone.}
\label{tab:sample_counts}
\begin{tabular}{@{}l r r r r@{}}
\toprule
\textbf{Client} & \textbf{Raw} & \textbf{Daylight} & \textbf{Model matrix} &
  \textbf{Matrix/daylight} \\
\midrule
S1 & 2{,}470 & 1{,}453 & 1{,}392 & 0.958 \\
S2 & 1{,}627 & 1{,}453 & 1{,}392 & 0.958 \\
S3 & 1{,}789 & 1{,}604 & 1{,}549 & 0.966 \\
S4 & 1{,}769 & 1{,}601 & 1{,}548 & 0.967 \\
S5 & 1{,}770 & 1{,}601 & 1{,}548 & 0.967 \\
S6 & 1{,}759 & 1{,}588 & 1{,}536 & 0.967 \\
S7 & 6{,}799 & 4{,}098 & 3{,}907 & 0.953 \\
\midrule
\textbf{Total} & \textbf{17{,}983} & \textbf{13{,}398} (74.5\%) &
  \textbf{12{,}872} (71.6\%) & --- \\
\bottomrule
\end{tabular}
\end{table}

\subsubsection{Lag Construction}

Lagged features were constructed by reindexing on $T-24\,\mathrm{h}$ rather
than using row offsets because the client timelines contained missing hours.
This ensured that the lagged capacity factor corresponded to the correct hour and
was set to \texttt{NaN} if no observation existed at that timestamp. The lag was computed on the client's
full timeline before applying the split filter. Validation rows could therefore include
lagged values from the training set. This procedure used earlier intervals rather than
future targets but does not establish literal dispatch-time availability. The feature standardizer was fitted only on the training
split to avoid information leakage from validation statistics.

\begin{table}[H]
\centering
\small
\setlength{\tabcolsep}{6pt}
\caption{Raw versus model-eligible counts by client and split, derived from
\texttt{split\_counts.csv} using Variant A and the past mask. Alternative B/common
rows describe the same observations and are not added again; B test rows are a
mask audit, not a model evaluation.}
\label{tab:retained_split_counts}
\begin{tabular*}{\textwidth}{@{\extracolsep{\fill}}lrrrrrr@{}}
\toprule
& \multicolumn{3}{c}{Raw rows} & \multicolumn{3}{c}{Model-eligible rows} \\
\cmidrule(lr){2-4}\cmidrule(l){5-7}
Client & Train & Val. & Test & Train & Val. & Test \\
\midrule
S1 & 1{,}729 & 370 & 371 & 882 & 247 & 263 \\
S2 & 1{,}138 & 244 & 245 & 934 & 229 & 229 \\
S3 & 1{,}252 & 268 & 269 & 1{,}051 & 246 & 252 \\
S4 & 1{,}238 & 265 & 266 & 1{,}051 & 248 & 249 \\
S5 & 1{,}239 & 265 & 266 & 1{,}051 & 248 & 249 \\
S6 & 1{,}231 & 264 & 264 & 1{,}041 & 248 & 247 \\
S7 & 4{,}759 & 1{,}020 & 1{,}020 & 2{,}709 & 593 & 605 \\
\midrule
Total & 12{,}586 & 2{,}696 & 2{,}701 & 8{,}719 & 2{,}059 & 2{,}094 \\
\bottomrule
\end{tabular*}
\end{table}

\subsection{Data Partitioning}
\begin{figure}
    \centering
    \includegraphics[width=\textwidth]{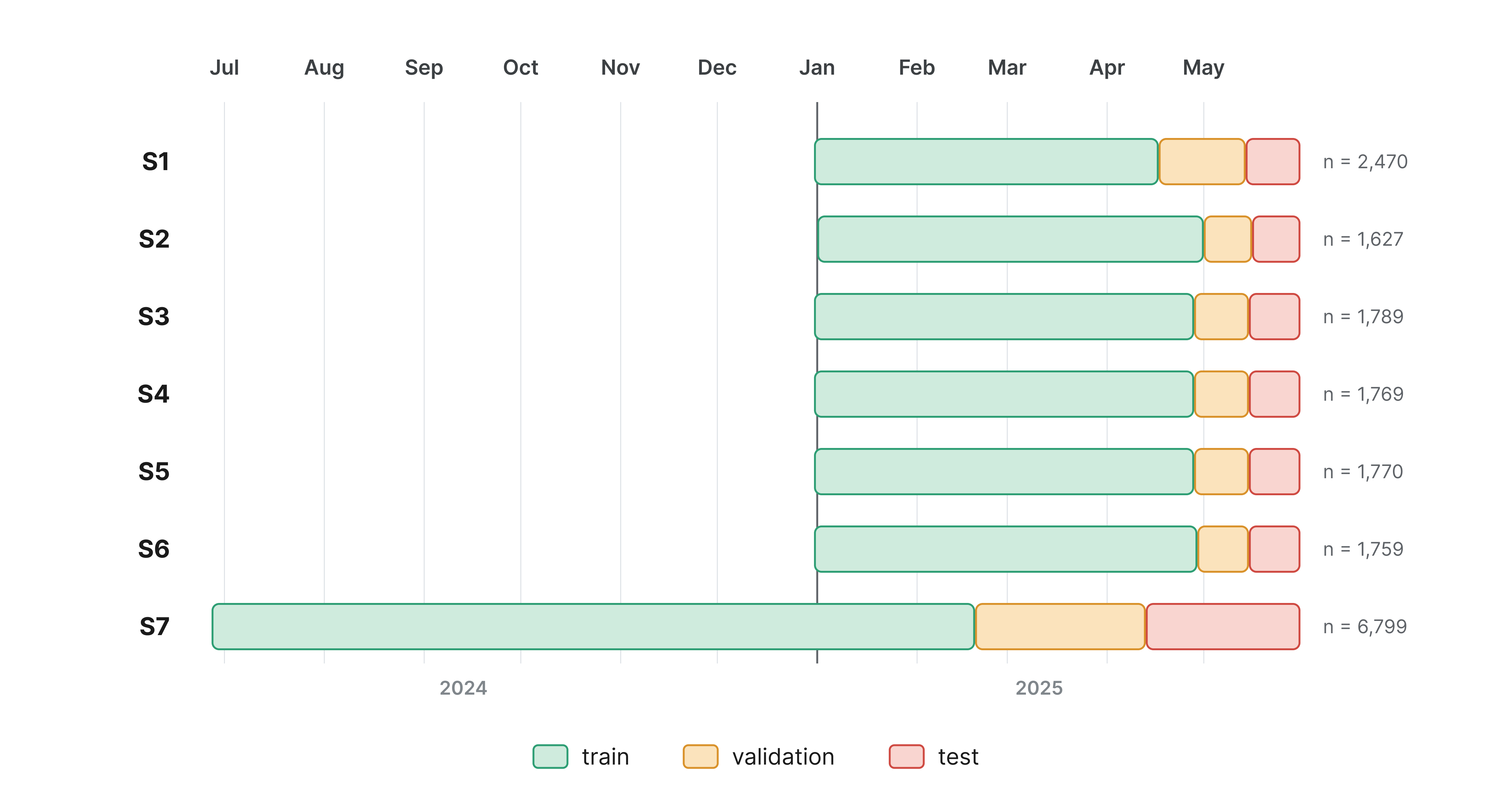}
    \caption{Chronological training, validation, and test split for each client (S1--S7).
    Boundaries and counts ($n$) refer to raw rows, not the retained model matrices.
    Table~\ref{tab:retained_split_counts} gives the latter.}
    \label{fig:per_client_split}
\end{figure}

Each client's raw observations were sorted chronologically and partitioned
independently by row position into training (70\%), validation (15\%), and test
(15\%) sets. The corresponding half-open intervals were
\[
[0,b_{\mathrm{train}}),\qquad
[b_{\mathrm{train}},b_{\mathrm{val}}),\qquad
[b_{\mathrm{val}},n).
\]
where $n$ is the client's total raw observation count and $b_{\mathrm{train}}$
and $b_{\mathrm{val}}$ mark the ends of the training and validation sets.
The deterministic split boundaries were frozen in a split manifest. When loading
the data, the feature builder verified that the total number of observations for
each client and the number in each split matched the manifest, raising an error
if any count differed.
Supplementary Table~\ref{tab:split_boundaries} reports the boundaries and split sizes, and
Figure~\ref{fig:per_client_split} shows the timing and relative duration of each
client's training, validation, and test periods.

The data were split separately for each client because the stations began
reporting at different times. A single cutoff date could therefore leave some stations
with too few training observations. The splits were based on the number of rows rather
than the length of the recording period, which maintained the intended split proportions
despite missing observations. Random splitting was not used because it could place
closely related observations from adjacent hours in different datasets.

The client-specific splits resulted in different seasonal coverage. Eligible
validation observations for the six shorter-running clients spanned
18 April--16 May 2025. For S7, they spanned 19 February--13 April 2025.

\subsection{Model Architecture}
The feed-forward MLP architecture was used for local, centralized, FedAvg,
and FedProx training. It comprised two fully connected hidden layers with 64 and
32 units, respectively, each followed by a rectified linear unit (ReLU)
activation, and a single linear output unit for capacity-factor regression.
Variant A used 12 past-only input features, whereas Variant B used 17 features,
including target-hour ERA5 reanalysis as a perfect-weather upper bound.
The corresponding models had 2{,}945 and 3{,}265 trainable parameters, respectively.
Supplementary Figure~\ref{fig:model_architecture} illustrates the architecture.

The neural models were trained using Adam with a learning rate of $10^{-3}$ and
mean squared error (MSE) loss. Centralized and federated early stopping used MAE
pooled across all validation observations. FedProx hyperparameter selection
used the same pooled validation set. This criterion gave equal weight to each
observation, so clients with more validation rows had greater influence on both
decisions. Early stopping used unclipped predictions and a patience of 25 epochs
or communication rounds of 300 at most. The weights with the lowest validation MAE were then
restored. FedProx hyperparameter selection used clipped predictions, and test
predictions were also clipped to $[0,1]$. All model and hyperparameter choices
were made on the validation set. The test set remained frozen until the protocol
had been fixed.

\subsection{Federated Learning Setup}
Federated training used a full-participation offline simulation with all seven
station clients. Before the first communication round, each client calculated
feature-wise sums, sums of squares, and sample counts from its training split.
The server aggregated these statistics to obtain global means and standard
deviations for standardizing the client datasets. This reproduced pooled
centralized scaling without pooling training feature rows for the scaling
calculation.

\begin{figure}
    \centering
    \includegraphics[width=\textwidth]{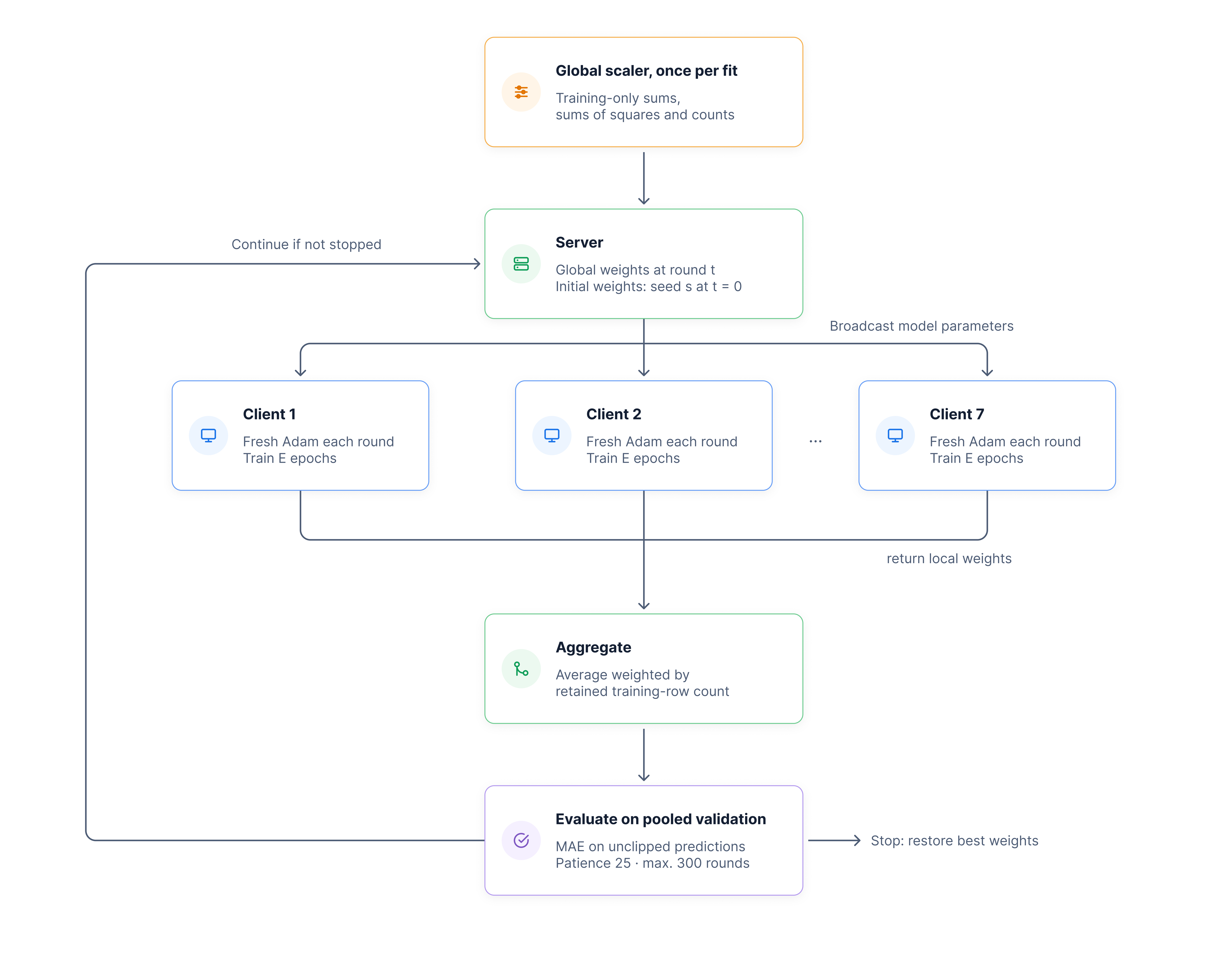}
    \caption{Full-participation simulated training. Seed $s$ initializes the model.
    Fresh client-round Adam updates are weighted by retained training-row count.
    The global scaler uses training-only statistics. Centrally concatenated validation
    predictors and targets support stopping on unclipped MAE. Best weights are restored
    at early stopping or the round limit. The host can access all arrays. Arrows do not
    represent enforced privacy boundaries. FedProx adds a squared-distance penalty to
    broadcast weights. Setting $\mu=0$ gives FedAvg.}
    \label{fig:federated_architecture}
\end{figure}

In each communication round, the server broadcast the global weights to all clients
(Figure~\ref{fig:federated_architecture}). Each client initialized a fresh Adam
optimizer, loaded the global weights, and trained on its standardized training
data for $E$ local epochs with a nominal batch size of 256. Shuffled indices were
partitioned using \path{array_split}, so actual batch sizes could differ from 256.
The server averaged the returned client weights in proportion to each client's
retained training-row count, rather than its raw split count. Training continued
for a maximum of 300 communication rounds and followed the pooled-validation
early-stopping and weight-restoration protocol described above.
Supplementary Algorithm~\ref{alg:fedavg_fedprox} summarizes the procedure.

FedProx used the same communication and aggregation procedure as FedAvg but added
a proximal penalty for deviation from the global weights received at the start of
each round; setting $\mu=0$ recovers FedAvg. Both methods were evaluated with
$E\in\{1,5\}$ local epochs to examine additional local computation between
aggregations. For each feature variant and value of $E$, FedProx was evaluated
with $\mu\in\{0.001,0.01,0.1,1.0\}$. Under Variant A, the clipped pooled
validation criterion selected $\mu=1$ at $E=1$, defining the primary federated
model for the frozen test.

Exploratory comparisons of model configurations on the validation set used
seed 0, and the past-only validation analysis was repeated with seeds 0--4.
For the frozen confirmatory test, the centralized MLP and selected FedProx model
were each trained with seeds
$0,1,2,3,$ and $4$, pairing the same seed across methods.

One local epoch per communication round ($E=1$) approximately matches data exposure
per centralized epoch, not computational cost. The architecture and primary
hyperparameters were held fixed across training regimes, but centralized Adam
retained its moment estimates across epochs, whereas federated Adam was
reinitialized for each client in every round. Aggregation and stopping dynamics
also differed (Supplementary Table~\ref{tab:optimization_protocol}), so performance differences
cannot be attributed solely to pooling versus federation. The $E=5$
configurations were secondary communication-computation analyses with more
client-side training per round. They were not compute-matched to centralized
training and were excluded from the primary non-inferiority conclusion.

\subsection{Evaluation Metrics}
Model performance was evaluated on valid daytime observations in the validation
and test sets using mean absolute error (MAE), root mean squared error (RMSE), and
an MAE-based skill score relative to persistence. MAE was the primary metric for
model evaluation and selection, while RMSE provided a complementary measure that
placed greater weight on large errors. Supplementary
Section~\ref{app:evaluation_metrics} provides the metric definitions and
aggregation formulas.

Metrics were reported separately for each client to show variation among stations.
Validation and test macro-averages gave equal weight to all seven clients,
regardless of their numbers of valid observations. The reported macro-averages
therefore describe performance for an equally weighted set of clients. This
aggregation differs from the pooled validation criterion used for training and
selection, which weights observations rather than clients equally.

\subsection{Confirmatory Test and Non-Inferiority Analysis}
\label{sec:noninferiority}
The confirmatory test compared the centralized MLP with full-participation
FedProx ($E=1$, $\mu=1$), using offline Variant A, which only included past PV and
weather data but excluded target-hour weather inputs.
For training seed $s$, the paired macro-MAE difference is defined in
Equation~\ref{eq:paired_seed_difference}:
\begin{equation}
D_s=\mathrm{MAE}_{\mathrm{FedProx},s}^{\mathrm{macro}}
    -\mathrm{MAE}_{\mathrm{centralized},s}^{\mathrm{macro}},
\label{eq:paired_seed_difference}
\end{equation}
The primary estimate, $D$, is the arithmetic mean of $D_s$ over seeds
$s\in\{0,1,2,3,4\}$; $D<0$ favors FedProx and $D>0$ favors centralized training.

The non-inferiority margin, $\Delta=0.005$ macro-MAE, was fixed before test
evaluation and recorded in the archived versioned decision log
\cite{PVForecastingRepository2026}. It represents a maximum acceptable increase of
0.5 percentage points in capacity-factor MAE. This margin is approximately 4.5\%
of the centralized model's macro-averaged validation MAE of 0.111986
($0.005/0.111986$). The tolerance was based on the
forecasting-error scale rather than economic cost because operational cost data
were unavailable. FedProx was declared non-inferior only if the upper limit of
the two-sided 95\% bootstrap confidence interval for the paired difference $D$
was below $0.005$.

The uncertainty analysis was specified before final test evaluation. It used a paired union-calendar day bootstrap with 10{,}000 repetitions and random seed 0. Predictions were paired by client, timestamp, and training seed. The union contained 49 dates present in at least one client's test set, while S1--S7 had 18, 16, 17, 17, 17, 17, and 49 test dates, respectively. Each repetition drew 49 dates with replacement and applied the same date weights to both methods, all clients, and all five training seeds. Each selected date contributed all available hourly observations for a client; dates without observations contributed no data, so the number of contributing date draws varied for clients with shorter calendars.

Within each repetition, MAE was recalculated for each method, client, and seed
using the selected observations. The FedProx-minus-centralized difference was
averaged equally across the seven clients and then across seeds 0--4. All
client--seed pairs had selected observations in every one of the 10,000
repetitions, as verified by reproducing the bootstrap analysis; no zero-observation
cases occurred. The two-sided 95\% percentile interval used the 2.5th and 97.5th
percentiles. The five fitted model pairs remained fixed, so the interval reflects
variation across sampled test days, not uncertainty from retraining on new
datasets or deployment in new geographical regions.

Two post hoc sensitivity analyses examined whether the non-inferiority conclusion
was robust to differences in test-date coverage across clients. The common-overlap
bootstrap used only the 16 dates shared by all seven clients and sampled 16 dates
with replacement, applying the same date weights across clients, methods, and seeds.
This approach ensured that every client contributed the same sampled dates, but it
excluded S7's additional dates and therefore evaluated a shorter period. The
client-stratified bootstrap instead retained each client's full set of test dates
and independently sampled the original number of dates available for that client.
Within each client, the sampled date weights were shared across methods and seeds,
which preserved method pairing and client-specific test periods but did not preserve
common sampled weather conditions across clients. Both sensitivity analyses used
10{,}000 repetitions, random seed 0, a two-sided 95\% percentile interval, and the
unchanged 0.005 non-inferiority margin.

%% file: sections/results.tex
\section{Results and Discussion}

\subsection{Exploratory Validation and Configuration Selection}

Validation screening compared the baselines and selected the federated configuration
before the test protocol was frozen. Under seed-0 Variant A, the local MLP ranked first
on all three metrics (Table~\ref{tab:validation_results}). It had a macro-MAE of 0.102,
an RMSE of 0.147, and a persistence skill of 0.157. The centralized MLP scored 0.112,
0.158, and 0.066, respectively. Ridge showed the reverse local--centralized ordering.
Centralized Ridge had lower MAE (0.115 versus 0.121), slightly lower RMSE (0.161 versus
0.163), and higher skill (0.047 versus 0.006). Pooling therefore coincided with lower
validation error for Ridge but higher error for the MLP. This reversal does not support
a model-independent effect of pooling.

\begin{table}[H]
\centering
\small
\renewcommand{\arraystretch}{1.10}
\caption{Validation performance in the initial seed-0 experiments. Values are
unweighted means across the seven clients. Lower MAE and RMSE are better.
Skill is calculated for each client as
$1-\mathrm{MAE}_{\mathrm{model}}/\mathrm{MAE}_{\mathrm{persistence}}$ and then
averaged across clients; higher values are better, and zero denotes equal MAE
to same-hour-yesterday persistence. A is the past-only offline variant. B is
the perfect-weather upper-bound variant, and its values are calculated from
the archived rounded client scores without implying additional precision or
a B test evaluation. $E$ is the number of local epochs per round. The selected
B FedProx configuration with $E=5$ uses $\mu=0.01$, whereas the corresponding
A configuration uses $\mu=1$.}
\label{tab:validation_results}
\begin{tabular*}{\textwidth}{@{\extracolsep{\fill}}llcccccc@{}}
\toprule
& & \multicolumn{3}{c}{\textbf{Variant A}} &
    \multicolumn{3}{c}{\textbf{Variant B}} \\
\cmidrule(lr){3-5}\cmidrule(lr){6-8}
\textbf{Model} & \textbf{A configuration} &
\textbf{MAE} & \textbf{RMSE} & \textbf{Skill} &
\textbf{MAE} & \textbf{RMSE} & \textbf{Skill} \\
\midrule
Persistence & ---                 & 0.121 & 0.189 & 0.000 & 0.121 & 0.189 & 0.000 \\
Ridge       & Local               & 0.121 & 0.163 & 0.006 & 0.095 & 0.132 & 0.222 \\
Ridge       & Centralized         & 0.115 & 0.161 & 0.047 & 0.097 & 0.139 & 0.196 \\
MLP         & Local               & 0.102 & 0.147 & 0.157 & 0.086 & 0.123 & 0.286 \\
MLP         & Centralized         & 0.112 & 0.158 & 0.066 & 0.090 & 0.135 & 0.260 \\
FedAvg      & $E=1$               & 0.119 & 0.168 & 0.007 & 0.098 & 0.144 & 0.188 \\
FedProx     & $E=1$, $\mu=1$      & 0.116 & 0.160 & 0.039 & 0.093 & 0.137 & 0.240 \\
FedAvg      & $E=5$               & 0.116 & 0.162 & 0.037 & 0.093 & 0.135 & 0.238 \\
FedProx     & $E=5$, $\mu=1$      & 0.114 & 0.158 & 0.057 & 0.093 & 0.135 & 0.239 \\
\bottomrule
\end{tabular*}
\end{table}

At $E=1$, FedProx had lower seed-0 validation error and higher skill than FedAvg.
Macro-MAE was 0.116 versus 0.119, RMSE was 0.160 versus 0.168, and persistence skill
was 0.039 versus 0.007. Both federated models nevertheless had higher MAE and lower
skill than the centralized MLP. Across validation seeds 0--4, mean macro-MAE was
0.111986 for the centralized MLP, 0.114235 for FedProx, and 0.116544 for FedAvg.
FedProx had lower macro-MAE than the centralized MLP in one of the five runs. FedProx
received a four-value search for $\mu$, whereas FedAvg had no matching search. These
comparisons are therefore exploratory.

FedProx $E=1$, $\mu=1$ was selected as the primary federated candidate. It was not the
best validation model overall. FedProx $E=5$ had a lower seed-0 Variant-A macro-MAE
than FedProx $E=1$ (0.114 versus 0.116), but used five local epochs per communication
round. The $E=1$ setting approximately matched the centralized model's data exposure,
not its computational cost. Within that setting, clipped pooled validation MAE selected
$\mu=1$ from the four evaluated values. These choices fixed the federated configuration
before the test without establishing validation superiority over the centralized MLP.

Variant B produced lower MAE and RMSE and higher persistence skill than Variant A for
every fitted model in Table~\ref{tab:validation_results}. The persistence baseline was
unchanged. For the centralized MLP, macro-MAE decreased from 0.112 to 0.090 and RMSE
from 0.158 to 0.135. Its skill increased from 0.066 to 0.260. For FedProx $E=1$,
macro-MAE decreased from 0.116 to 0.093 and RMSE from 0.160 to 0.137. Its skill
increased from 0.039 to 0.240. Variant B includes target-hour ERA5 reanalysis that was
unavailable at the nominal $T-24$ forecast origin. It therefore provides a
validation-only, non-operational perfect-weather upper bound rather than a deployable
forecasting configuration.

\subsection{Primary Non-Inferiority Finding}

\begin{figure}[t]
    \centering
    \includegraphics[width=\textwidth]{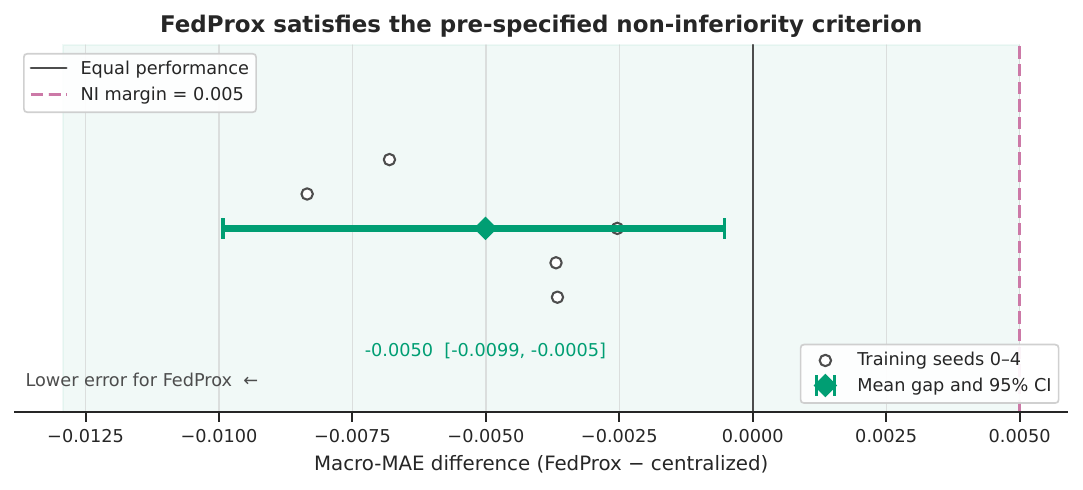}
    \caption{Frozen primary non-inferiority analysis. The estimand is the macro-MAE
    difference between FedProx and centralized training, averaged over training seeds
    0--4; negative values favor FedProx. The diamond and horizontal interval show the
    observed gap and paired 49-date union-calendar day bootstrap 95\% confidence interval,
    while circles show the five seed-specific gaps. The vertical line at zero denotes
    equal observed error and the dashed line at $\Delta=0.005$ is the pre-specified
    non-inferiority margin. Non-inferiority is concluded only when the confidence
    interval's upper endpoint is strictly below the margin, as occurred here. This
    confirmatory analysis uses past-only offline Variant A and its future-independent past
    row-set mask.}
    \label{fig:test_noninferiority}
\end{figure}

The selected FedProx configuration was next compared with the centralized MLP on the
Variant-A test set. The local models, Ridge models, FedAvg, $E=5$ settings, and
Variant B remained exploratory validation comparisons. Across five paired training
seeds, the centralized MLP had a macro-MAE of 0.133405. FedProx $E=1$, $\mu=1$ had a
macro-MAE of 0.128397. The pre-specified difference, $D$,
was $-0.005008$. Its paired union-calendar day bootstrap 95\% confidence interval was
$[-0.009926,-0.000537]$ (Figure~\ref{fig:test_noninferiority}). The upper endpoint,
$-0.000537$, was below the fixed margin of 0.005. FedProx therefore met the primary
non-inferiority criterion.

The gap of 0.005008 capacity-factor units equals 0.5008 percentage points and a 3.75\%
lower macro-MAE relative to the centralized MLP. The margin of $\Delta=0.005$ allowed
a maximum increase of 0.5 percentage points. This limit was about 4.5\% of the
centralized validation macro-MAE of 0.111986. It was based on the error scale because
the study did not include an operational cost model. A post hoc check varied $\Delta$
from 0 to 0.010, with no change in the decision
(Supplementary Table~\ref{tab:margin_sensitivity}).

The primary analysis used the union calendar. Two post hoc checks then tested the effect
of unequal client test dates (Table~\ref{tab:calendar_sensitivity}). The primary and
client-stratified designs retained all available dates and the gap of $-0.005008$. The
common-overlap design used only the 16 dates shared by all clients. Its gap was
$-0.006766$ for that shorter period. The upper 95\% confidence limits were $-0.000537$,
$-0.001945$, and $-0.002050$ for the primary, overlap, and stratified designs,
respectively. Each limit was below 0.005, so the decision held under all three calendar
treatments. The overlap design, however, estimated performance over a different period.

\begin{table}[H]
\centering
\small
\setlength{\tabcolsep}{4pt}
\renewcommand{\arraystretch}{1.12}
\caption{Calendar-bootstrap sensitivity of the non-inferiority conclusion. The
union-calendar design is the originally specified primary analysis; the other designs
are post hoc. $D$ is FedProx minus centralized macro-MAE. All analyses used 10{,}000
repetitions, random seed 0, a two-sided 95\% percentile interval, and $\Delta=0.005$;
NI denotes the non-inferiority decision.}
\label{tab:calendar_sensitivity}
\begin{tabularx}{\textwidth}{@{}>{\raggedright\arraybackslash}Xrrrrc@{}}
\toprule
\textbf{Design} & \shortstack{\textbf{Centralized}\\\textbf{MAE}} &
\shortstack{\textbf{FedProx}\\\textbf{MAE}} &
\textbf{$D$} & \textbf{95\% CI} & \textbf{NI} \\
\midrule
Union calendar (primary)\newline
49 dates; synchronized & 0.133405 & 0.128397 &
$-0.005008$ & $[-0.009926,-0.000537]$ & Pass \\
\addlinespace[2pt]
Common overlap (post hoc)\newline
16 shared dates; synchronized & 0.139345 & 0.132578 &
$-0.006766$ & $[-0.012120,-0.001945]$ & Pass \\
\addlinespace[2pt]
Client stratified (post hoc)\newline
16--49 dates/client; independent & 0.133405 & 0.128397 &
$-0.005008$ & $[-0.008097,-0.002050]$ & Pass \\
\bottomrule
\end{tabularx}
\end{table}

All primary and calendar-check intervals lay below zero. The study did not pre-specify a
superiority test, so this direction does not establish a formal superiority claim. The
experiment also did not identify why the methods differed. It did not isolate the
proximal term, and the two training regimes differed in optimizer states, model averaging,
and stopping rules. The gap therefore cannot be assigned solely to proximal
regularization or to federation.

The finding applies only to the seven clients, past-only Variant A, the selected
architecture and training budgets, and the frozen test procedure. It does not establish
privacy or dispatch-time availability. The simulation host could access all station
arrays, validation data were joined centrally, and complete lagged intervals became
available after their interval-start labels. The retrospective capacity rule was also
retained without a new sensitivity test.

\subsection{Robustness Across Training Seeds}

The primary estimate averaged five paired training seeds. Table~\ref{tab:test_seed_results}
shows the result from each fit. Every gap was negative, ranging from $-0.008349$ to
$-0.002538$ macro-MAE. Thus, each seed produced lower macro-MAE for FedProx than for the
centralized MLP. Seeds 3 and 4 produced the largest gaps, $-0.008349$ and $-0.006807$.
At those seeds, centralized macro-MAE rose to 0.136819 and 0.136788, while FedProx
remained at 0.128470 and 0.129981. The larger gaps therefore arose mainly from higher
centralized errors rather than unusually low FedProx errors.

\begin{table}[H]
\centering
\small
\setlength{\tabcolsep}{8pt}
\caption{Frozen test macro-MAE by paired training seed. Macro-MAE is the unweighted mean
of the seven client MAEs. $D$ is FedProx minus centralized, so negative values favor
FedProx. The mean row is the primary point estimate; seeds are repeated fits on the same
test observations, not independent test datasets. Differences were calculated before
rounding, so displayed method values may not subtract exactly to the reported $D$.}
\label{tab:test_seed_results}
\begin{tabular}{@{}rrrr@{}}
\toprule
\textbf{Seed} & \textbf{Centralized MLP} & \textbf{FedProx} & \textbf{$D$} \\
\midrule
0 & 0.130968 & 0.127309 & -0.003660 \\
1 & 0.130857 & 0.127171 & -0.003686 \\
2 & 0.131592 & 0.129054 & -0.002538 \\
3 & 0.136819 & 0.128470 & -0.008349 \\
4 & 0.136788 & 0.129981 & -0.006807 \\
\midrule
Mean & 0.133405 & 0.128397 & -0.005008 \\
\bottomrule
\end{tabular}
\end{table}

Using five pre-specified seeds reduced reliance on one fitted model and prevented the
selection of a favorable seed after test evaluation. However, every model pair used the
same station-hour observations. The five rows therefore do not represent independent
test datasets. They assess robustness to model initialization only. Formal uncertainty
still comes from the paired union-calendar day bootstrap used for the primary confidence
interval.

\subsection{Client-Level Performance and Heterogeneity}

\begin{figure}[t]
    \centering
    \includegraphics[width=0.92\textwidth]{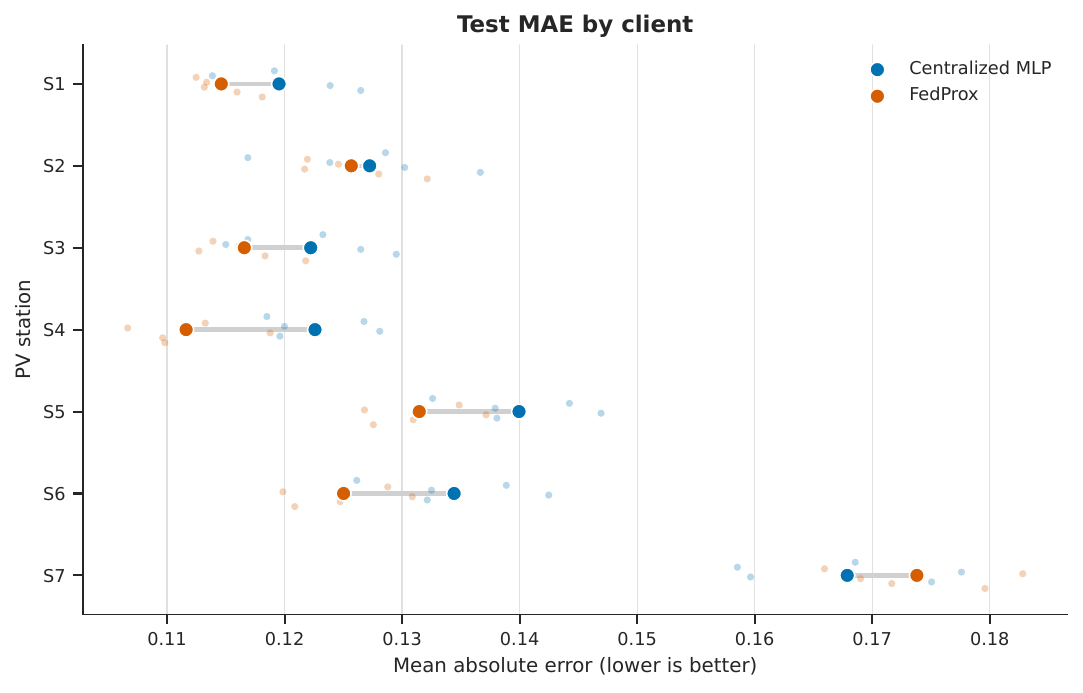}
    \caption{Frozen test MAE by client. Large markers show the mean across training seeds
    0--4 and smaller markers show individual-seed values for the centralized MLP and
    FedProx $E=1$, $\mu=1$; gray segments connect the two method means within each
    station. Lower MAE is better. FedProx had lower mean MAE at S1--S6, whereas the
    centralized MLP had lower mean MAE at S7. Results use past-only offline Variant A
    and its future-independent past row-set mask.}
    \label{fig:test_client_mae}
\end{figure}

The consistent seed-level direction did not imply uniform performance across clients
(Table~\ref{tab:test_client_metrics} and Figure~\ref{fig:test_client_mae}). FedProx had
lower mean MAE at S1--S6. After rounding, the gaps ranged from 0.002 at S2 to 0.011 at
S4. S7 showed the opposite pattern: its centralized MAE was lower by 0.006. RMSE followed
the same station-level ordering, with lower FedProx values at S1--S6 and a slightly lower
centralized value at S7. The macro result therefore describes the average across clients,
not a uniform station-level difference.

\begin{table}[H]
\centering
\small
\renewcommand{\arraystretch}{1.10}
\caption{Frozen test metrics by station, reported as means across the five training seeds.
$n_{\mathrm{test}}$ is constant across seeds and methods. C denotes centralized MLP and
FP denotes FedProx $E=1$, $\mu=1$. $D_{\mathrm{MAE}}=\mathrm{MAE}_{\mathrm{FP}}-
\mathrm{MAE}_{\mathrm{C}}$, so negative differences favor FedProx. Lower MAE and RMSE
are better; positive skill indicates lower MAE than same-hour-yesterday persistence.
Differences were calculated from unrounded means, so displayed MAEs may not subtract
exactly to the reported $D_{\mathrm{MAE}}$.}
\label{tab:test_client_metrics}
\begin{tabular*}{\textwidth}{@{\extracolsep{\fill}}lrrrrrrrr@{}}
\toprule
\textbf{Station} & \textbf{$n_{\mathrm{test}}$} & \textbf{C MAE} & \textbf{FP MAE} &
\textbf{$D_{\mathrm{MAE}}$} & \textbf{C RMSE} & \textbf{FP RMSE} &
\textbf{C skill} & \textbf{FP skill} \\
\midrule
S1 & 263 & 0.120 & 0.115 & -0.005 & 0.177 & 0.161 & 0.085 & 0.123 \\
S2 & 229 & 0.127 & 0.126 & -0.002 & 0.187 & 0.174 & 0.039 & 0.050 \\
S3 & 252 & 0.122 & 0.117 & -0.006 & 0.183 & 0.165 & 0.053 & 0.097 \\
S4 & 249 & 0.123 & 0.112 & -0.011 & 0.175 & 0.156 & -0.003 & 0.087 \\
S5 & 249 & 0.140 & 0.131 & -0.008 & 0.187 & 0.172 & 0.107 & 0.161 \\
S6 & 247 & 0.134 & 0.125 & -0.009 & 0.179 & 0.164 & 0.098 & 0.161 \\
S7 & 605 & 0.168 & 0.174 &  0.006 & 0.230 & 0.233 & -0.119 & -0.159 \\
\bottomrule
\end{tabular*}
\end{table}

Persistence skill further qualified the station-level pattern. At S4, the centralized
MLP was marginally worse than persistence, with a mean skill of $-0.003$, while FedProx
had a positive skill of 0.087. At S7, neither method outperformed persistence. The
centralized and FedProx skill scores were $-0.119$ and $-0.159$, indicating MAEs 11.9\%
and 15.9\% above persistence, respectively. S7 also had the largest test set, with 605
observations. It did not dominate macro-MAE because each of the seven clients received
equal weight. These results show that the favorable macro difference did not extend to
every client or guarantee improvement over persistence.

Additional exploratory seed-0 validation comparisons of local epochs and proximal
regularization, and their communication-computation limitations, are provided in
Supplementary Section~\ref{supp:secondary_fl}.

\subsection{Practical Implications}

The non-inferiority finding supports further evaluation of federated forecasting where PV operators cannot centrally pool operational records. In such settings, exchanging model updates could enable collaborative training while retaining raw records locally. The present experiment addresses the predictive-performance component of this potential application: the selected FedProx configuration met the pre-specified tolerance relative to centralized training. However, the simulation used centrally pooled validation data and did not establish enforced data isolation or privacy protection.

Forecasts can inform the energy commitments of PV operators and aggregators, with deviations between committed and delivered generation affecting imbalance exposure. Previous research connects forecast uncertainty to commitment strategies and shows that economic outcomes depend on deviation penalties and decision rules \cite{BASHIR2021100477}. In PV-plus-storage systems, forecasts also inform when and how much the battery should charge or discharge, subject to operational constraints \cite{YANG20192326}. Comparisons of dispatch strategies show that sensitivity to forecast errors varies with the controller \cite{MirletzLaws2023}. Predictive non-inferiority therefore does not establish equivalent revenues, balancing requirements, or battery performance.

Practical adoption also requires station-level assessment. FedProx reduced mean MAE relative to centralized training at six stations, but both methods underperformed persistence at S7. This variation reinforces evidence that the benefits of federated forecasting differ among participants \cite{Nachtigall2024}. Operational evaluation should therefore compare federated, centralized, local, and persistence forecasts within the same commitment or battery-scheduling framework, using inputs available at forecast issuance. The present offline comparison, based on a 24-hour timestamp lag, provides a basis for that evaluation rather than evidence of realized operational benefits.

%% file: sections/conclusion.tex
\section{Conclusion}

The study evaluates whether selected FedProx training is
non-inferior to centralized MLP training for PV capacity-factor forecasting
in an offline comparison using a 24-hour timestamp lag.
Across five paired training seeds, FedProx with $E=1$ and
$\mu=1$ achieved a macro-MAE of 0.128397, compared to 0.133405 for
centralized training, giving a mean paired difference of $-0.005008$
with a 95\% union-calendar day bootstrap confidence interval of
$[-0.009926, -0.000537]$.
Because the upper bound was below the pre-specified non-inferiority
margin of 0.005, FedProx satisfied the non-inferiority criterion.
Post hoc calendar sensitivity analyses did not change this conclusion.
All five seed-specific differences favored FedProx although station-level
performance was heterogeneous.
FedProx reduced mean MAE at six stations, whereas
centralized training performed better at S7, where both methods
underperformed persistence.

These findings support further evaluation of federated
forecasting for collaboration among PV operators where operational
records cannot be centrally pooled. Such forecasts could inform energy
commitments and BESS scheduling, although predictive non-inferiority
does not establish operational or economic benefits.
The simulation used centrally pooled validation data
and did not demonstrate privacy protection or forecast-input
availability at issuance.

The findings are limited to one observation period and to seven
geographically close stations in North Macedonia, with full client
participation.
Client-specific temporal splits, retrospective capacity
definitions, and differences in optimization further qualify the
comparison.
Future work should evaluate generalization across longer
periods and additional regions, alongside station-level suitability,
privacy mechanisms, and operational performance within market-commitment
or battery-scheduling frameworks.

%% file: sections/experiment_scope_table.tex
\begin{table}[H]
\centering
\small
\renewcommand{\arraystretch}{1.12}
\caption{Evidence scope, from \texttt{experiment\_scope.csv}. Persistence is the
skill reference, not a separate inferential endpoint. No additional model is fitted or
selected on the test split. A and B denote past-only and perfect-weather inputs.}
\label{tab:experiment_scope}
\begin{tabularx}{\textwidth}{@{}>{\raggedright\arraybackslash}p{2.4cm}>{\raggedright\arraybackslash}p{1.25cm}>{\raggedright\arraybackslash}p{1.1cm}>{\raggedright\arraybackslash}X@{}}
\toprule
Stage & Seeds & Variant & Models and role \\
\midrule
Exploratory validation & 0 & A, B & Local/centralized Ridge and MLP;
FedAvg/FedProx, $E=1,5$; configuration selection and A/B comparison. \\
\addlinespace
Validation robustness & 0--4 & A & Centralized MLP, FedAvg $E=1$, and
FedProx $E=1$; initialization robustness, not test inference. \\
\addlinespace
Confirmatory test & 0--4 paired & A & Centralized MLP versus selected FedProx
$E=1$, $\mu=1$ only; frozen non-inferiority analysis. \\
\bottomrule
\end{tabularx}
\end{table}

%% file: sections/feature_dimensions_table.tex
\begin{table}[H]
\centering
\small
\caption{Matrix widths and actual MLP parameter counts from
\texttt{feature\_dimensions.csv}. The retained constant column is
\texttt{is\_daylight}. Counts of pooled nonconstant columns are identical across
training, validation, and test matrices; they are not matrix ranks. B test-side
counts are descriptive only. Both networks use 64 and 32 ReLU hidden units and one
linear output.}
\label{tab:feature_dimensions}
\begin{tabular*}{\textwidth}{@{\extracolsep{\fill}}lrrr@{}}
\toprule
Variant & Input width & Nonconstant columns & Parameters \\
\midrule
A & 12 & 11 & 2{,}945 \\
B & 17 & 16 & 3{,}265 \\
\bottomrule
\end{tabular*}
\end{table}

%% file: sections/optimization_table.tex
\begin{table}[H]
\centering
\small
\renewcommand{\arraystretch}{1.1}
\caption{Optimization protocol from \texttt{optimization\_protocol.csv}. Shared
settings are Adam, learning rate $10^{-3}$, nominal batch size 256 using
\texttt{array\_split}, and strict improvement in pooled validation MAE on unclipped
predictions. Best weights are restored at either early stopping or the budget limit.
$E=1$ approximately matches data exposure, not exact compute.}
\label{tab:optimization_protocol}
\begin{tabularx}{\textwidth}{@{}>{\raggedright\arraybackslash}p{2.3cm}>{\raggedright\arraybackslash}X>{\raggedright\arraybackslash}X@{}}
\toprule
Setting & Centralized MLP & FedAvg / FedProx \\
\midrule
Objective & MSE & MSE; FedProx adds $\frac{\mu}{2}\|\mathbf{w}_k-\mathbf{w}^{t}\|_2^2$ \\
Optimizer state & Retained across epochs & Fresh per client-round \\
Budget / patience & 300 epochs / 25 epochs & 300 rounds / 25 rounds \\
Aggregation & None & Retained training-row-count weighted weights \\
Initialization & \texttt{torch.manual\_seed(s)} & \texttt{torch.manual\_seed(s)} \\
Shuffle stream & \texttt{default\_rng(s)} & Persistent \texttt{default\_rng(s+i)} for zero-based ordered client $i$ \\
Stopping data & Pooled validation predictors and targets & Centrally concatenated validation predictors and targets \\
Test scoring & Predictions clipped to $[0,1]$ & Predictions clipped to $[0,1]$ \\
\bottomrule
\end{tabularx}
\end{table}

%% file: sections/appendix.tex
\section{Non-Inferiority Margin Sensitivity}
\label{supp:margin_sensitivity}

Table~\ref{tab:margin_sensitivity} reports a post-hoc sensitivity check of the
non-inferiority decision across alternative margins. The primary margin
$\Delta=0.005$ was pre-specified; all other margins were evaluated post hoc and do not
replace the primary analysis.

\begin{table}[H]
\centering
\small
\setlength{\tabcolsep}{7pt}
\renewcommand{\arraystretch}{1.12}
\caption{Sensitivity of the non-inferiority decision to the margin. The upper confidence
interval endpoint is unchanged because the fitted models, predictions, and bootstrap
analysis for the primary union-calendar design are unchanged. The margin $\Delta=0.005$ is the pre-specified primary margin;
all other margins are post-hoc sensitivity checks.}
\label{tab:margin_sensitivity}
\begin{tabular}{@{}rrrrl@{}}
\toprule
$\Delta$ & \textbf{CF percentage points} & \textbf{Upper 95\% CI} &
\textbf{Non-inferiority} & \textbf{Role} \\
\midrule
0       & 0    & -0.000537 & Pass & Post-hoc sensitivity \\
0.0025  & 0.25 & -0.000537 & Pass & Post-hoc sensitivity \\
0.005   & 0.5  & -0.000537 & Pass & Pre-specified primary \\
0.0075  & 0.75 & -0.000537 & Pass & Post-hoc sensitivity \\
0.010   & 1.0  & -0.000537 & Pass & Post-hoc sensitivity \\
\bottomrule
\end{tabular}
\end{table}

\section{Evaluation Metric Definitions}
\label{app:evaluation_metrics}

Let $y_{k,i}$ and $\hat{y}_{k,i}$ denote the observed and predicted capacity factors
for observation $i$ of client $k$, and let $n_k$ be the number of evaluated
observations for that client. MAE measures the average absolute forecasting error
in capacity-factor units:
\begin{equation}
\mathrm{MAE}_k =
\frac{1}{n_k}\sum_{i=1}^{n_k}
\left|y_{k,i}-\hat{y}_{k,i}\right|.
\label{eq:mae}
\end{equation}
Because capacity factor is dimensionless and normalized by station capacity,
MAE values are directly comparable across clients of different sizes. RMSE is
defined as
\begin{equation}
\mathrm{RMSE}_k =
\sqrt{\frac{1}{n_k}\sum_{i=1}^{n_k}
\left(y_{k,i}-\hat{y}_{k,i}\right)^2}.
\label{eq:rmse}
\end{equation}
Squaring the residuals gives greater weight to large errors, so RMSE complements
MAE by indicating whether forecasts occasionally have particularly large errors.
Lower values of both metrics indicate better performance.

For target hour $T$, the 24-hour persistence forecast is
\begin{equation}
\hat{y}^{\mathrm{pers}}_{k,T}=y_{k,T-24\,\mathrm{h}}.
\label{eq:persistence}
\end{equation}
The skill score is based on MAE rather than RMSE:
\begin{equation}
\mathrm{Skill}_k =
1-\frac{\mathrm{MAE}_{k,\mathrm{model}}}
        {\mathrm{MAE}_{k,\mathrm{pers}}}.
\label{eq:skill}
\end{equation}
Positive, zero, and negative skill scores indicate better, equal, and worse
performance than persistence, respectively. A score of $0.20$ corresponds to a
20\% reduction in MAE. Persistence has a skill score of zero by construction.

For a client-level metric $q_k$, the macro-average is
\begin{equation}
q_{\mathrm{macro}}=\frac{1}{K}\sum_{k=1}^{K}q_k,
\label{eq:macro_metric}
\end{equation}
Each of the $K=7$ clients contributes equally, regardless of its evaluation-set size.
Pooled MAE is defined as
\begin{equation}
\mathrm{MAE}_{\mathrm{pooled}}=
\frac{1}{\sum_{k=1}^{K}n_k}
\sum_{k=1}^{K}\sum_{i=1}^{n_k}
\left|y_{k,i}-\hat{y}_{k,i}\right|,
\label{eq:pooled_mae}
\end{equation}
This formulation weights observations equally and therefore gives greater
influence to clients with larger evaluation sets.